\documentclass[10pt,twocolumn,letterpaper]{article}

\usepackage{ijcb}
\usepackage{hyperref}

\usepackage{times}
\usepackage{epsfig}
\usepackage{graphicx}
\usepackage{amsmath}
\usepackage{amssymb}
\usepackage{mathtools}
\usepackage{fancyhdr}
\usepackage{url}
\usepackage{enumitem}
\usepackage{subcaption}
\usepackage{multirow}
\usepackage{tabularx}
\usepackage{xspace}
\usepackage[normalem]{ulem}
\useunder{\uline}{\ul}{}
\usepackage[norule]{footmisc}

\setlist[itemize]{leftmargin=*}

\ijcbfinalcopy % *** Uncomment this line for the final submission

\ifijcbfinal\fi

\makeatletter
\def\ps@IEEEtitlepagestyle{
\def\@oddfoot{\mycopyrightnotice}
\def\@evenfoot{}
}
\def\mycopyrightnotice{
{\hfill \footnotesize
\vspace{-10mm}
978-1-6654-3780-6/21/\$31.00 \copyright 2021 IEEE\hfill}
}
\makeatother

\begin{document}

%%%%%%%%% TITLE
\title{MFR 2021: Masked Face Recognition Competition}

{
\author{ Fadi Boutros$^{1,2,*}$,
Naser Damer$^{1,2,*}$,
Jan Niklas Kolf$^{1,*}$,
Kiran Raja $^{3,*}$, \\
Florian Kirchbuchner$^{1,*}$, Raghavendra Ramachandra$^{3,*}$,
Arjan Kuijper$^{1,2,*}$, \\
Pengcheng Fang$^{4,+}$,
Chao Zhang$^{4,+}$,
Fei Wang$^{4,+}$,
David Montero$^{5,6,+}$,
Naiara Aginako$^{6,+}$,\\
Basilio Sierra$^{6,+}$,
Marcos Nieto$^{5,+}$,
Mustafa Ekrem Erak{\i}n$^{7,+}$,
U\u{g}ur Demir$^{7,+}$, 
Haz{\i}m Kemal Ekenel$^{7,+}$,\\
Asaki Kataoka$^{8,+}$,
Kohei Ichikawa$^{8,+}$, 
Shizuma Kubo$^{8,+}$,
Jie Zhang$^{9,10,+}$,
Mingjie He$^{9,10,+}$, \\
Dan Han$^{9,10,+}$,  
Shiguang Shan$^{9,10,+}$,  
Klemen Grm$^{11,+}$,
Vitomir \v{S}truc$^{11,+}$,
Sachith Seneviratne$^{12,+}$, \\
Nuran Kasthuriarachchi$^{13,+}$, 
Sanka Rasnayaka$^{14,+}$,
Pedro C. Neto$^{15,16,+}$,
Ana F. Sequeira$^{16,+}$, \\
Joao Ribeiro Pinto$^{15,16,+}$, 
Mohsen Saffari$^{15,16,+}$,
Jaime S. Cardoso$^{15,16,+}$ 
\\
$^{1}$Fraunhofer Institute for Computer Graphics Research IGD, Germany -
$^{2}$TU Darmstadt, Germany\\
$^{3}$Norwegian University of Science and Technology, Norway - 
$^{4}$TYAI, China\\
$^{5}$VICOMTECH, Spain - 
$^{6}$University of the Basque Country, Spain \\ 
$^{7}$ SiMiT Lab, Istanbul Technical University, Turkey 
$^{8}$ACES, Inc, Japan \\
$^{9}$Institute of Computing Technology, Chinese Academy of Sciences, China \\
$^{10}$University of Chinese Academy of Sciences, China\\
$^{11}$University of Ljubljana, Slovenia - 
$^{12}$University of Melbourne, Australia \\ 
$^{13}$University of Moratuwa, Sri Lanka -
$^{14}$National University of Singapore, Singapore \\
$^{15}$INESC TEC, Portugal -
$^{16}$University of Porto, Portugal\\
$^{*}$Competition organizer. $^{+}$Competition participant.\\
Email: {fadi.boutros@igd.fraunhofer.de}
\vspace{-4mm}
}
}

\maketitle
\thispagestyle{empty}

%%%%%%%%% ABSTRACT
\vspace{-2mm}
\begin{abstract}
\vspace{-2mm}
This paper presents a summary of the Masked Face Recognition Competitions (MFR) held within the 2021 International Joint Conference on Biometrics (IJCB 2021).
The competition attracted a total of 10 participating teams with valid submissions.
The affiliations of these teams are diverse and associated with academia and industry in nine different countries.
These teams successfully submitted 18 valid solutions.
The competition is designed to motivate solutions aiming at enhancing the face recognition accuracy of masked faces.
Moreover, the competition considered the deployability of the proposed solutions by taking the compactness of the face recognition models into account.
A private dataset representing a collaborative, multi-session, real masked, capture scenario is used to evaluate the submitted solutions.
In comparison to one of the top-performing academic face recognition solutions, 10 out of the 18 submitted solutions did score higher masked face verification accuracy.
\end{abstract}

{\let\thefootnote\relax\footnotetext{\mycopyrightnotice}}

\vspace{-3mm}
\section{Introduction}
\label{sec:int}
\vspace{-2mm}

Given the current COVID-19 pandemic, it is essential to enable contactless and smooth-running operations, especially in contact-sensitive facilities like airports. With the ever-enhancing performance of face recognition, the technology has been preferred as a contactless means of verifying identities in applications ranging from border control to logical access control on consumer electronics. However, wearing masks is now essential to prevent the spread of contagious diseases and has been currently forced in public places in many countries. The performance, and thus the trust in contactless identity verification through face recognition can be impacted by the presence of a mask \cite{DBLP:journals/corr/abs-2102-09258}. The effect of wearing a mask on face recognition in a collaborative environment is currently a sensitive issue.
This competition is the first to attract and present technical solutions that enhance the accuracy of masked face recognition on real face masks and in a collaborative  verification scenario.

In a recent study, the National Institute of Standards and Technology (NIST), as a part of the ongoing Face Recognition Vendor Test (FRVT), has published a specific study (FRVT -Part 6A) on the effect of face masks on the performance on face recognition systems provided by vendors \cite{ngan2020ongoing}. The NIST study concluded that the algorithm accuracy with masked faces declined substantially. One of the main study limitations is the use of simulated masked images under the questioned assumption that their effect represents that of real face masks.
The Department of Homeland Security has conducted an evaluation with similar goals, however on more realistic data \cite{DHS-Masks-2020}. They also concluded with the significant negative effect of wearing masks on the accuracy of automatic face recognition solutions.
A study by Damer et al. \cite{DBLP:conf/biosig/DamerGCBKK20} evaluated the verification performance drop in 3 face biometric systems when verifying masked vs not-masked faces, in comparison to verifying not-masked faces to each other. The authors presented limited data (24 subjects), however, with real masks and multiple capture sessions. They concluded by noting the bigger effect of masks on genuine pairs decisions, in comparison to imposter pairs decisions. This study has been extended \cite{https://doi.org/10.1049/bme2.12040} with a larger database and evaluation on both synthetic and real masks, pointing out the questionable use of simulated masks to represent the real mask effect on face recognition. Recent work has evaluated the human performance in recognizing masked faces, in comparison to automatic face recognition solutions \cite{DBLP:journals/corr/abs-2103-01924}. The study concluded with a set of take-home messages that pointed to the correlated effect of wearing masks on both, human recognizers and automatic face recognition. Beyond recognition, facial masks showed to affect both, the vulnerability of face recognition to presentations attacks, and the detectability of these attacks \cite{DBLP:journals/corr/abs-2103-01546}.

%what work have been done to fix it

There were only a few works that address enhancing the recognition performance of masked faces.
Li et al. \cite{li2021cropping} proposed to use an attention-based method to train a face recognition model on the periocular area of masked faces. This presented improvement in the masked face recognition performance, however in a limited evaluation. Moreover, the proposed approach essentially only maps the problem into a periocular recognition problem.
A recent preprint by \cite{anwar2020masked} presented a relatively small dataset of 53 identities crawled from the internet. The work proposed to fine-tune FacenNet model \cite{DBLP:conf/cvpr/SchroffKP15} using simulated masked face images to improve the recognition accuracy. Wang et al. \cite{wang2020masked} presented three datasets crawled from the internet for face recognition, detection, and simulated masked faces.
%The face recognition dataset contains 5000 masked face images of 525 identities and 90000 unmasked face images of the same 525 identities.
The authors claim to improve the verification accuracy from 50\% to 95\% on masked faces. However, they did not provide any information about the evaluation protocol, proposed solution, or implementation details.
Moreover, the published part of the dataset does not contain pairs of not-masked vs masked images for the majority of identities. 
A work by Montero et al. \cite{DBLP:journals/corr/abs-2104-09874} proposed to combine ArcFace loss with a specially designed mask-usage classification loss to enhance masked face recognition performance. 
Boutros et al. \cite{DBLP:journals/corr/abs-2103-01716} proposed a template unmasking approach that can be adapted on the top of any face recognition network. This approach aims to create unmasked-like templates from masked faces. This goal was achieved on top of multiple networks by the proposed self-restrained triplet loss \cite{DBLP:journals/corr/abs-2103-01716}.
On a related matter, a rapid number of works are published to address the detection of wearing a face mask \cite{app11052070,loey2021hybrid,wang2021wearmask,qin2020identifying}. These studies did not address the effect of wearing a mask on the performance of face recognition or present solution to improve masked face recognition.

Besides the exclusive interest in face recognition accuracy, there is a growing interest in compact face recognition models \cite{martinez2021benchmarking}. This interest is driven by the demand for face recognition deployment on consumer devices and the need to enhance the throughput of face recognition processes.
A major challenge has been organized in ICCV 2019 to motivate researchers to build lightweight face recognition models \cite{DBLP:conf/iccvw/DengGZDLS19}.
MobileFaceNets are an example of such face recognition models \cite{DBLP:conf/ccbr/ChenLGH18}. MixedFaceNets \cite{mixedfacenets} are a recent example where mixed depthwise convolutional kernels, with a tailored head and embedding design and a shuffle operation, are utilized to achieve high recognition accuracies with extremely light models.

Motivated by (a) the hygiene-driven wide use of facial masks, (b) the proven performance decay of existing face recognition solutions when processing masked faces, (c) the need to motivate novel research in the direction of enhancing masked face recognition accuracy, and (d) the requirement of light-weight models by various applications, we conducted the IJCB Masked Face Recognition Competition 2021 (IJCB-MFR-2021). The competition attracted submissions from academic and industry teams with a wide international representation. The final participation toll was 10 teams with valid submissions. These teams submitted 18 valid solutions.
The solutions were evaluated on a database collected to represent a collaborative face verification scenario with individuals wearing real face masks. 
This paper summarises this competition with a detailed presentation of the submitted solutions and the achieved results in terms of masked vs masked face verification accuracy, masked vs not-masked face verification accuracy, and the compactness of the recognition models.

In the next sections, we start by introducing the competition evaluation database, the evaluation criteria, and the participating teams. Then, in Section \ref{sec:alg}, short descriptions of the submitted solutions are listed. In Section \ref{sec:res}, we present and discuss the achieved results along with listing the winning submissions. We end the paper in Section \ref{sec:con} with a final general conclusion.

\vspace{-3mm}
\section{Database, evaluation criteria, and participants}
\label{sec:data}
\vspace{-1mm}
\subsection{Database}
\vspace{-1mm}
The evaluation data, the masked face recognition competition data (MFRC-21), simulates a collaborative, yet varying scenario. Such as the situation in automatic border control gates or unlocking personal devices with face recognition, where the mask, illumination, and background can change. The database is collected by the hosting institute and not available publicly. The data is collected on three different, not necessarily consecutive days. We consider each of these days as one session. On each day, the subjects have collected three videos, each of a minimum length of 5 seconds (used as single image frames). 
The videos are collected from static webcams (not handheld), while the subjects are requested to look at the camera, simulating a login scenario.
The data is collected by subjects at their residences during the pandemic-induced home-office period.
The first session is considered a reference session, while the other two were considered probe sessions. Each day contained three types of captures, no mask, masked with natural illumination, masked with additional illumination. The database participants were asked to remove eyeglasses only when the frame is considered very thick. No other restrictions were imposed, such as background or mask type and its consistency over days, to simulate realistic scenarios.
The first second of each video was neglected to avoid possible biases related to the subject interaction with the capture device.
After the neglected one second, three seconds were considered.
From these three seconds, 10 frames are extracted with a gap of 9 frames between each consecutive frame, knowing that all videos are captured at a frame rate of 30 frames per second.

\begin{figure*}[h]
     \centering
     \begin{subfigure}[b]{0.48\textwidth}
         \centering
         \includegraphics[width=0.5\textwidth]{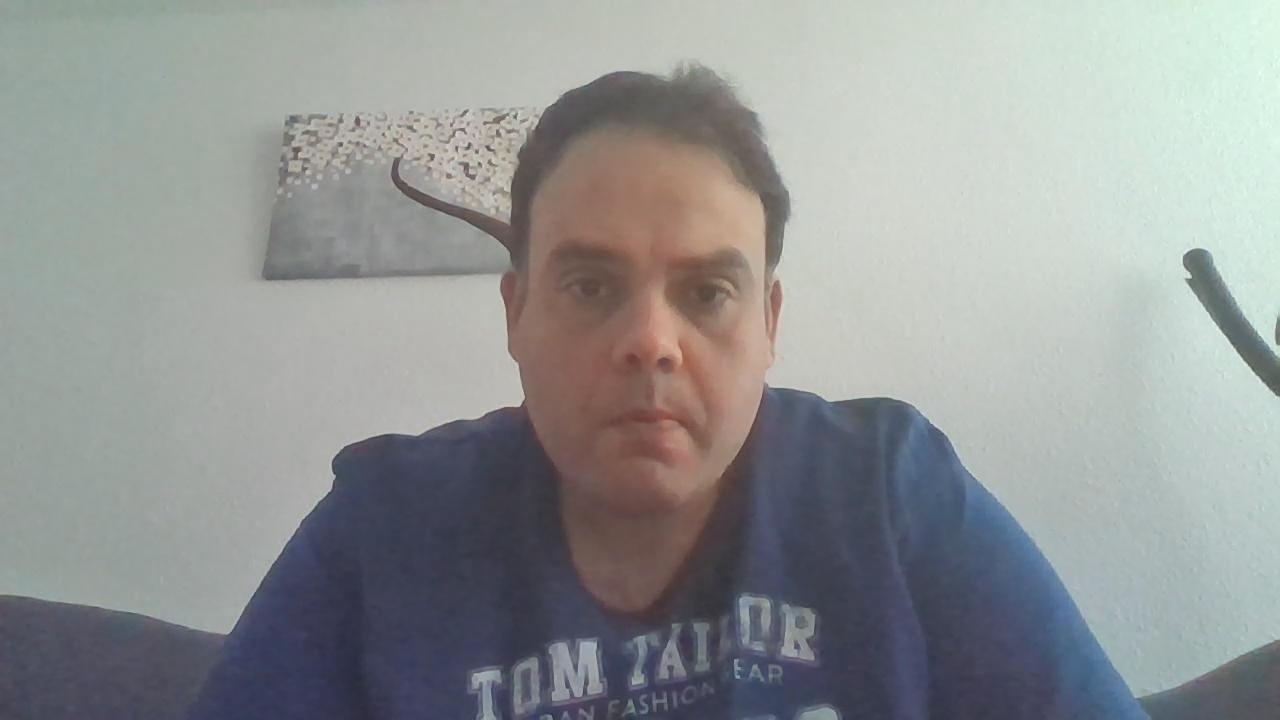}\includegraphics[width=0.5\textwidth]{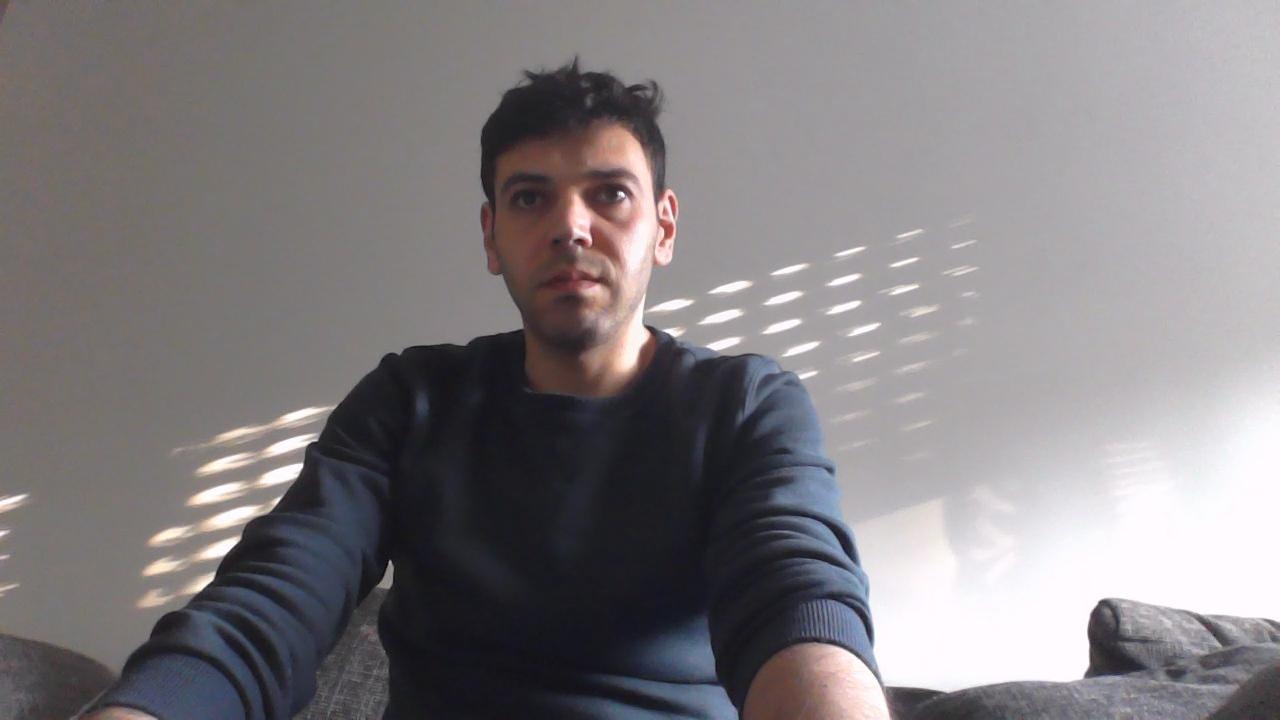}
         \caption{Not-masked baseline faces (BLR)}
         \label{fig:samp:BL}
     \end{subfigure}
     \begin{subfigure}[b]{0.48\textwidth}
         \centering
         \includegraphics[width=0.5\textwidth]{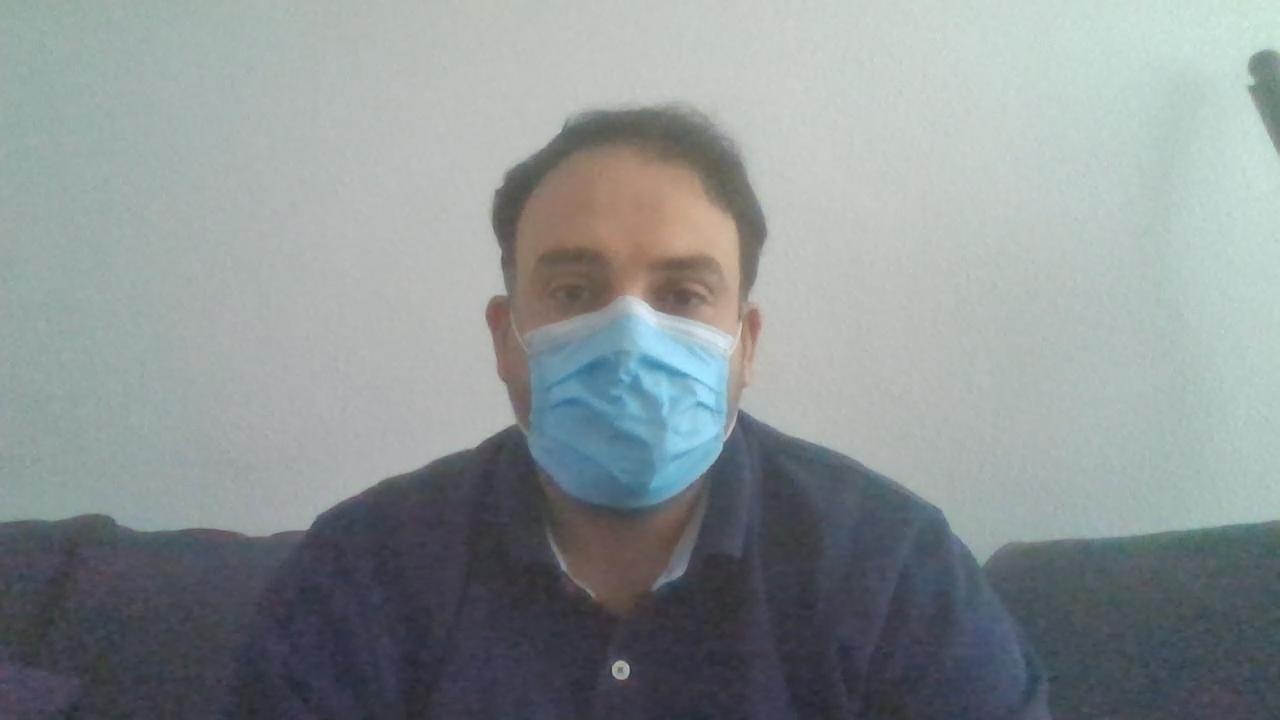}\includegraphics[width=0.5\textwidth]{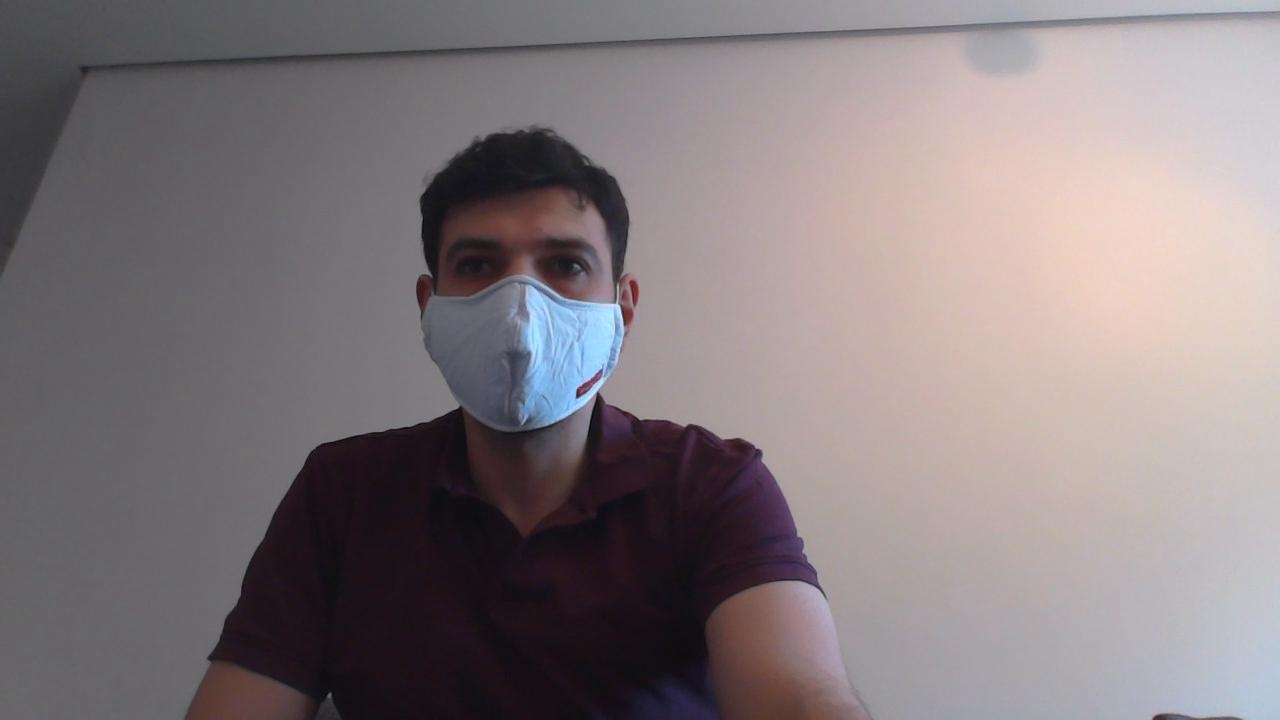}
         \caption{Masked faces (MR/MP)}
         \label{fig:sam:M}
     \end{subfigure}
     \vspace{-2mm}
      \caption{Samples of the MFRC-21 database from the two capture types (BLR and  MR/MP). MR and MP have  similar capture settings, MR on the first setting and MP on the second and third session.}
        \label{fig:samples}
        \vspace{-4mm}
\end{figure*}

The final considered portions of the database in the competition are (a) the not-masked baseline reference from the first session (noted as BLR), (b) the masked reference from the first session (noted as MR), and (c) the masked face probes from the second and third sessions under both illumination scenarios (noted as MP). A summary of the used database is presented in Table \ref{tab:DB} and samples of the database are presented in Figure \ref{fig:samples}. The database contained 47 subjects, all of them participated in all the sessions. All the subject provided their informed consent to use the data for research purposes.

Two evaluation setups are considered, (a) not-masked vs masked, where all images in BLR are compared to all images in MP (noted as BLR-MP), and (b) masked vs. masked, where all images in MR are compared to all images in MP (noted as MR-MP).

\begin{table}[]
\begin{tabular}{|l|l|l|l|l|}
\hline
Session            & \multicolumn{3}{l|}{\begin{tabular}[c]{@{}l@{}}Session 1: \\ References\end{tabular}} & \begin{tabular}[c]{@{}l@{}}Session 2 and 3:\\ Probes\end{tabular} \\ \hline
Data split         & BLR                           & \multicolumn{2}{l|}{MR}                               & MP                                                                \\ \hline
Number of Captures & 470                           & \multicolumn{2}{l|}{940}                              & 1880                                                              \\ \hline
\end{tabular}
\caption{An overview of the MFRC-21 database structure.}
\label{tab:DB}
\vspace{-6mm}
\end{table}

\subsection{Evaluation criteria}
\label{sec:eval_criteria}
%The baseline performance evaluation will be based on the open-source implementation of the ArcFace model. The considered model architecture is LResNet100E-IR trained on ms1m-refine-v2 database with ArcFace loss function. The pre-trained model is available on the official ArcFace Github repository.

The solutions evaluation will be based on both, the verification performance and the compactness of the used mode/models. The verification evaluation will be based on the verification performance of masked vs. not-masked verification pairs, as this is the common scenario, where the reference is not-masked, while the probe is masked, e.g. in entry to a secure access area. This scenario will be noted as BLR-MP. However, the performance of masked vs. masked verification pairs is also be reported in this paper. This scenario is noted as MR-MP.

The verification performance is evaluated and reported as the false non-match rate (FNMR) at different operation points FMR100, FMR1000, which are the lowest FNMR for a false match rate (FMR) $<1.0\%$ and $<0.1\%$, respectively.  The verification performance evaluation of the submitted solutions is based on FMR100.
To get an indication of generalizability, we also report a separability measure between the genuine and imposter comparison scores. This is measured by the Fisher Discriminant Ratio (FDR) as formulaed in \cite{poh2004study}. 

To consider the deployability of the participating solutions, we will also consider the compactness of the model (represented by the number of trainable parameters \cite{DBLP:journals/corr/abs-2103-06877}) in the final ranking. The participants are asked to report the number of trainable parameters and can be asked to provide their solutions to validate this number.

The final teams ranking is be based on a weighted Borda count, where the participants will be ranked by (a) the verification metric as mentioned above (noted as Rank-a), and (b) by the number of trainable parameters in their model/models (notes as Rank-b). For Rank-a, the solutions with lower FMR100 are ranked first, and for Rank-b, the solutions with the lower number of trainable parameters are ranked first.
In the final ranking, Rank-a will have 75\% weight and Rank-b will have 25\% weight. 
Each participant is given a Borda count (BC) for each ranking criteria (BC-a and BC-b).
For example, if solution X is ranked first out of 10 participants in the verification performance rank-a (BC-a =9) and third out of 10 solutions in model compactness Rank-b (BC-b = 7) (this corresponds to BC = total number of solutions – rank). Then the weighted Borda count w-BC = 0.75x9+0.25x7= 8.5. Therefore, the final score of solution X is 8.5 and higher indicates a better solution. The solutions are ranked from the highest w-BC to the lowest w-BC.

\subsection{Submission and evaluation process}

Each of the teams was requested to submit their solutions as Win32 or Linux console applications.
These applications should be able to accept three parameters, evaluation-list (text file), landmarks (text file), and an output path.
The evaluation-list 
contains pairs of the path to the reference and probe images and a label for each of the compared images, indicating if the image is masked or not.
The landmarks provided a bounding box and five landmark locations of the images as detected by the MTCNN solution \cite{zhang2016joint}.
Only the pairs of images with valid detected faces are provided to the solutions in the evaluation-list.
From the initial considered data, the face detector \cite{zhang2016joint} did not provide valid face detections.
For the BLR-MP pairs, 4.42\% of the pairs contained invalid detections of faces and thus not considered in the evaluation. for the MR-MP pairs, 4.75\% of the pairs contained invalid detections of faces and thus not considered in the evaluation.
The output of the solution application script is a text file containing comparison scores for each pair in the evaluation-list.

\subsection{Competition participants}
The competition aimed at attracting participants with a high geographic and activity variation. The call for participation was shared on the International Joint Conference on Biometrics (IJCB 2021) website, on the competition own website \footnote{https://sites.google.com/view/ijcb-mfr-2021/home}, on public computer vision mailing lists (e.g. CVML e-Mailing List), and through private e-Mailing lists.
The call for participation has attracted 12 registered teams. Out of these, 10 teams have submitted valid solutions. These 10 teams have affiliations based in nine different countries. Seven of the 10 teams are affiliated with academic institutions, two are affiliated with the industry, and one team has both academic and industry affiliations.
Only one of the participating teams has chosen to be anonymous.
Each team was allowed to submit up to two solutions. 
The total number of validly submitted solutions is 18.
A summary of the participating teams is presented in Table \ref{tab:team}.

\begin{table*}
\centering

\resizebox{\textwidth}{!}{%
\begin{tabular}{|l|l|l|l|} 
\hline
Solution      & Team members                                                                                                                                             & Affiliations                                                                                                                                                     & Type of institution        \\ 
\hline
A1\_Simple    & Asaki Kataoka, Kohei Ichikawa, Shizuma Kubo                                                                                                      & ACES, Inc, Japan                                                                                                                                                        & Industry                   \\ 
\hline
TYAI          & Pengcheng Fang, Chao Zhang, Fei Wang                                                                                                             & TYAI, China                                                                                                                                                             & Industry                   \\ 
\hline
MaskedArcFace & \multirow{2}{*}{\begin{tabular}[c]{@{}l@{}}David Montero, Naiara Aginako Basilio Sierra,\\  Marcos Nieto\end{tabular}}                            & \multirow{2}{*}{Vicomtech, Spain - University of the Basque Country, Spain}                                                                                              & \multirow{2}{*}{Academic}  \\ 
\cline{1-1}
MTArcFace     &                                                                                                                                                     &                                                                                                                                                                  &                            \\ 
\hline
MFR-NMRE-F    & \multirow{2}{*}{Klemen Grm, Vitomir \v{S}truc}                                                                                                      & \multirow{2}{*}{University of Ljubljana, Slovenia}                                                                                                                         & \multirow{2}{*}{Academic}  \\ 
\cline{1-1}
MFR-NMRE-B    &                                                                                                                                                     &                                                                                                                                                                  &                            \\ 
\hline
MUFM Net      & \multirow{2}{*}{\begin{tabular}[c]{@{}l@{}}Sachith Seneviratne, Nuran Kasthuriarachchi, \\  Sanka Rasnayaka\end{tabular}}                         & \multirow{2}{*}{\begin{tabular}[c]{@{}l@{}} University of Melbourne, Australia - National University of Singapore, \\ Singapore -  University of Moratuwa, Sri Lanka\end{tabular}}                      & \multirow{2}{*}{Academic}  \\ 
\cline{1-1}
EMUFM Net     &                                                                                                                                                     &                                                                                                                                                                  &                            \\ 
\hline
VIPLFACE-M    & \multirow{2}{*}{Jie Zhang , Mingjie He, Dan Han, Shiguang Shan}                                                                                                 & \multirow{2}{*}{\begin{tabular}[c]{@{}l@{}}Institute of Computing Technology, Chinese Academy of Sciences, China, \\ University of Chinese Academy of Sciences, China\end{tabular}}                 & \multirow{2}{*}{Academic}  \\ 
\cline{1-1}
VIPLFACE-G    &                                                                                                                                                     &                                                                                                                                                                  &                            \\ 
\hline
SMT-MFR-1     & \multirow{2}{*}{\begin{tabular}[c]{@{}l@{}}Mustafa Ekrem Erak{\i}n, U\u{g}ur Demir, \\ Haz{\i}m Kemal Ekenel\end{tabular}}                                 & \multirow{2}{*}{\begin{tabular}[c]{@{}l@{}}Smart Interaction and Machine Intelligence Lab (SiMiT Lab), Istanbul \\ Technical University, Turkey\end{tabular}}                         & \multirow{2}{*}{Academic}  \\ 
\cline{1-1}
SMT-MFR-2     &                                                                                                                                                     &                                                                                                                                                                  &                            \\ 
\hline
LMI-SMT-MFR-1 & \multirow{2}{*}{\begin{tabular}[c]{@{}l@{}}Mustafa Ekrem Erak{\i}n, U\u{g}ur Demir, \\ Haz{\i}m Kemal Ekenel, Klemen Grm, Vitomir \v{S}truc\end{tabular}}     & \multirow{2}{*}{\begin{tabular}[c]{@{}l@{}} Istanbul Technical University, Turkey -  University of Ljubljana, Slovenia\end{tabular}}                        & \multirow{2}{*}{Academic}  \\ 
\cline{1-1}
LMI-SMT-MFR-2 &                                                                                                                                                     &                                                                                                                                                                  &                            \\ 
\hline
IM-MFR        & \multirow{2}{*}{\begin{tabular}[c]{@{}l@{}}Pedro C. Neto, Ana F. Sequeira, João Ribeiro Pinto, \\ Mohsen Saffari, Jaime S. Cardoso\end{tabular}} & \multirow{2}{*}{\begin{tabular}[c]{@{}l@{}}INESC TEC, Portugal -  University of Porto, Faculty  of Engineering \\ (FEUP), Portugal\end{tabular}} & \multirow{2}{*}{Academic}          \\ 
\cline{1-1}
IM-AMFR       &                                                                                                                                                     &                                                                                                                                                                  &                            \\ 
\hline
Anonymous-1 & \multirow{2}{*}{Anonymous}                                                                                                                          & \multirow{2}{*}{Anonymous}                                                                                                                                       & \multirow{2}{*}{mix}  \\ 
\cline{1-1}
Anonymous-2 &                                                                                                                                                     &                                                                                                                                                                  &                            \\
\hline
\end{tabular}}
\vspace{-2mm}
\caption{A summary of the submitted solutions,
participant team members, affiliations, and type of institutions (Industry, Academic, or mix). The table lists the abbreviations of each submitted solution. Details of the submitted algorithms are in Section \ref{sec:alg}. }
\label{tab:team}
\vspace{-2mm}

\end{table*}

% \usepackage{multirow}
% \usepackage{graphicx}
\begin{table*}[]
\centering
\resizebox{\textwidth}{!}{%
\begin{tabular}{|l|l|l|l|l|l|l|l|l|l|l|}
\hline
\multirow{2}{*}{Solution} & \multicolumn{5}{l|}{Verification performance} & \multicolumn{3}{l|}{Compactness} & \multicolumn{2}{l|}{Joint} \\ \cline{2-11} 
              & FMR100 & FMR1000 & FDR     & Rank-a & BC-a & number of parameters & Rank-b & BC-b & w-BC  & Rank       \\ \hline  \hline
Baseline      & 0.06009 & 0.07154 & 8.6899  & -  & -  &  65155648 & -  & -  & -     & -          \\ \hline  \hline
TYAI          & 0.05095 & 0.05503  & 11.2005 & 1      & 17   & 70737600             & 14     & 4    & 13.75 & \textbf{1} \\ \hline
MaskedArcFace & 0.05687 & 0.05963  & 10.4484 & 5      & 13   & 43589824             & 6      & 12   & 12.75 & \textbf{2} \\ \hline
SMT-MFR-2     & 0.05584 & 0.06268  & 11.2025 & 3      & 15   & 65131000             & 12     & 6    & 12.75 & \textbf{2} \\ \hline
A1\_Simple    & 0.05538 & 0.06113  & 8.5147  & 2      & 16   & 87389138             & 16     & 2    & 12.5  & \textbf{3} \\ \hline
VIPLFACE-M    & 0.05681 & 0.06279  & 8.2371  & 4      & 14   & 65128768             & 10     & 8    & 12.5  & \textbf{3} \\ \hline
MTArcFace     & 0.05699 & 0.05860  & 10.7497 & 6      & 12   & 43640002             & 7      & 11   & 11.75 & 4          \\ \hline
SMT-MFR-1     & 0.05704 & 0.06003  & 10.6824 & 7      & 11   & 65131000             & 12     & 6    & 9.75  & 5          \\ \hline
VIPLFACE-G    & 0.05750 & 0.07269  & 8.1693  & 9      & 9    & 65128768             & 10     & 8    & 8.75  & 6          \\ \hline
MFR-NMRE-B    & 0.05819 & 0.08344  & 7.9504  & 10     & 8    & 43723943             & 8      & 10   & 8.5   & 7          \\ \hline
LMI-SMT-MFR-1 & 0.05722 & 0.06205  & 9.7384  & 8      & 10   & 108854000            & 17     & 1    & 7.75  & 8          \\ \hline
MFR-NMRE-F    & 0.08125 & 0.17660  & 5.3876  & 12     & 6    & 43723943             & 8      & 10   & 7     & 9          \\ \hline
MUFM Net      & 0.17579 & 0.40489  & 4.4640  & 14     & 4    & 25636712             & 3      & 15   & 6.75  & 10         \\ \hline
IM-AMFR       & 0.28252 & 0.47608  & 3.7414  & 15     & 3    & 36898792             & 4      & 14   & 5.75  & 11         \\ \hline
LMI-SMT-MFR-2 & 0.05848 & 0.07096  & 8.5278  & 11     & 7    & 108854000            & 17     & 1    & 5.5   & 12         \\ \hline
Anonymous-1   & 0.92536 & 0.96596  & 0.1011  & 17     & 1    & 23777281             & 1      & 17   & 5     & 13         \\ \hline
IM-MFR        & 0.28447 & 0.47430  & 3.7369  & 16     & 2    & 36898792             & 4      & 14   & 5     & 13         \\ \hline
EMUFM Net     & 0.16239 & 0.35681  & 4.5445  & 13     & 5    & 76910136             & 15     & 3    & 4.5   & 14         \\ \hline
Anonymous-2   & 0.97125 & 0.99517  & 0.0426  & 18     & 0    & 23777281             & 1      & 17   & 4.25  & 15         \\ \hline
\end{tabular}%
}
\vspace{-2mm}
\caption{The comparative evaluation of the submitted solutions on the MFRC-21 dataset. The results are presented in terms of verification performance including FMR100, FMR1000, and FDR, and the model compactness in terms of the number of trainable parameters. The FMR100 and FMR1000 are given as absolute values.
The rank of the verification performance (Rank-a) is based on FMR100 and the rank of the solution compactness (Rank-b) is based on the number of parameters. Rank-a has 75\% weight and Rank-b has 25\% weight.
The results are ordered based on weighted Borda count (w-BC). }
\label{tab:blr_mp}
\vspace{-4mm}
\end{table*}

% \usepackage{multirow}
% \usepackage{graphicx}
\begin{table}[]
\centering
\resizebox{\linewidth}{!}{%
\begin{tabular}{|l|l|l|l|l|}
\hline
\multirow{2}{*}{Solution} & \multicolumn{4}{l|}{Verification performance} \\ \cline{2-5} 
                          & FMR100    & FMR1000   & FDR        & Rank   \\ \hline  \hline
Baseline                  & 0.05925   & 0.06504   & 9.68640     & -      \\ \hline  \hline
TYAI                      & 0.04489    & 0.05961    & 12.36306   & 1      \\ \hline
VIPLFACE-M                & 0.05759    & 0.06788    & 8.98593    & 2      \\ \hline
A1\_Simple                & 0.05771    & 0.06368    & 10.48611   & 3      \\ \hline
SMT-MFR-2                 & 0.05792    & 0.06172    & 11.30901   & 4      \\ \hline
MaskedArcFace             & 0.05825    & 0.06245    & 10.57307   & 5      \\ \hline
SMT-MFR-1                 & 0.05825    & 0.06012    & 11.03444   & 6      \\ \hline
VIPLFACE-G                & 0.05843    & 0.06359    & 9.41466    & 7      \\ \hline
MTArcFace                 & 0.0585     & 0.06390     & 10.16996   & 8      \\ \hline
LMI-SMT-MFR-1             & 0.05856    & 0.06061    & 9.90914    & 9      \\ \hline
LMI-SMT-MFR-2             & 0.05916    & 0.06586    & 8.87424    & 10     \\ \hline
MFR-NMRE-B                & 0.05970     & 0.12903    & 8.11963    & 11     \\ \hline
MFR-NMRE-F                & 0.09630     & 0.1989     & 4.73224    & 12     \\ \hline
EMUFM Net                 & 0.15045    & 0.31945    & 4.45317    & 13     \\ \hline
MUFM Net                  & 0.16354    & 0.37607    & 4.43278    & 14     \\ \hline
IM-AMFR                   & 0.23507    & 0.40265    & 3.94744    & 15     \\ \hline
IM-MFR                    & 0.23661    & 0.40373    & 3.94905    & 16     \\ \hline
Anonymous-1               & 0.89481    & 0.97584    & 0.19968    & 17     \\ \hline
Anonymous-2               & 0.9114     & 0.98102    & 0.16569    & 18     \\ \hline
\end{tabular}%
}
\vspace{-1mm}
\caption{The comparative evaluation results of the submitted solutions. The verification evaluation is based on the verification performance of masked vs. masked verification pairs where references and probes are masked. The performances are reported in terms of FMR-100, FMR-1000 and FDR. The FMR100 and FMR1000 are given as absolute values. The reported results are ordered based on FMR-100.}
\label{tab:mr_mp}
\end{table}

% \usepackage{multirow}
% \usepackage{graphicx}
\begin{table}[]
\centering
\resizebox{\linewidth}{!}{%
\begin{tabular}{|l|l|l|l|l|l|}
\hline
Solution &
  Input size &
  FM &
  Loss function &
  RM & % Real masked data (training
  SM \\ \hline %  Synthetically masked data (training) 
\textbf{A1\_Simple}                     & 112 x 112 & 512  & ArcFace            & No  & Yes \\ \hline
\textbf{TYAI}                           & 112 x 112 & 512  & Sub-center ArcFace & No  & Yes \\ \hline
\textbf{MaskedArcFace}                  & 112 x 112 & 512  & ArcFace            & No  & Yes \\ \hline
MTArcFace                      & 112 x 112 & 512  & ArcFace            & No  & Yes \\ \hline
MFR-NMRE-F                     & 96 x 192  & 2048 & CE      & No  & No  \\ \hline
MFR-NMRE-B                     & 112 x 224 & 2048 & CE      & No  & No  \\ \hline
MUFM Net                       & 224 x 224 & 2048 & CE      & No  & Yes \\ \hline
EMUFM Net                      & 224 x 224 & 2048 & CE      & No  & Yes \\ \hline
\textbf{VIPLFACE-M}                    & 112 x 112 & 512  & ArcFace            & No  & Yes \\ \hline
VIPLFACE-G                     & 112 x 112 & 512  & ArcFace            & No  & No  \\ \hline
SMT-MFR-1                      & 112 x 112 & 512  & ArcFace            & Yes & No  \\ \hline
\textbf{SMT-MFR-2}                      & 112 x 112 & 512  & ArcFace            & Yes & No  \\ \hline
\multirow{2}{*}{LMI-SMT-MFR-1} & 96 x 192  & 2048 & CE      & No  & No  \\ \cline{2-6} 
                               & 112 x 112 & 512  & ArcFace            & Yes & No  \\ \hline
\multirow{2}{*}{LMI-SMT-MFR-2} & 112 x 224 & 2048 & CE      & No  & No  \\ \cline{2-6} 
                               & 112 x 112 & 512  & ArcFace            & Yes & No  \\ \hline
IM-MFR &
  224 x 224 &
  512 &
  \begin{tabular}[c]{@{}l@{}}CE, triplet loss\\ and MSE\end{tabular} &
  No &
  Yes \\ \hline
IM-AMFR &
  224 x 224 &
  512 &
  \begin{tabular}[c]{@{}l@{}}CE, triplet loss\\ and MSE\end{tabular} &
  No &
  Yes \\ \hline
Anonymous-1                    & 160 x 160 & 512  & CE      & No  & Yes \\ \hline
Anonymous-2                    & 160 x 160 & 512  & CE      & No  & Yes \\ \hline
\end{tabular}%
}
\vspace{-1mm}
\caption{Basic details of the submitted solutions including, the input image size, the feature embedding size (FM), the loss function used for training, the use of real masked faces (RM), and simulated masked faces (SM) in the training process. The solutions in bold are the ones ranked top in the competition. Note that all the top-ranked solutions used a version of the ArcFace loss \cite{DBLP:conf/cvpr/DengGXZ19,DBLP:conf/eccv/DengGLGZ20}.}
\label{tab:solutions_summary}
\end{table}

\section{Submitted solutions}
\label{sec:alg}

Ten teams have been registered for MFR 2021 competition and submitted 18 valid solutions. Table \ref{tab:team} presents a summary of the registered team members and their affiliation, submitted solutions, and type of institution of each registered team (Academic, Industry, or mix of both academic and industry). In the following, we provide a brief description of the valid submitted solutions: 

\vspace{-2mm}
\paragraph{A1\_Simple} employed ArcFace \cite{DBLP:conf/cvpr/DengGXZ19} to train a ResNet model. A1\_Simple applied MaskTheFace \cite{anwar2020masked} method to synthetically generate masked face images in the training dataset- MS1MV2. A1\_Simple is trained with cosine annealing LR scheduling to adjust the learning rate. 
In the evaluation phase, A1\_Simple used the provided landmark facial point and bounding box in the MFRC-21 to align and crop the face image to $112 \times 112$. The feature embedding of the presented solution is of size 512-D. The model is trained with ArcFace loss. During the training phase, three data augmentation methods are used- random resized crops, random horizontal flip, and color jittering.
\vspace{-2mm}
\paragraph{TYAI} solution uses Sub-center ArcFace \cite{DBLP:conf/eccv/DengGLGZ20} and ir-ResNet152 model to train a masked face recognition model on Glint360K dataset \cite{DBLP:journals/corr/abs-2010-05222}. The proposed solution randomly augmented half of the training dataset with a synthetic generated mask using five types of transparent masks. The input image size of the proposed model is $112 \times 112$ and the size of the output feature embedding is 512-D. During the training, additional four data augmentation methods are used: random crop by resizing the image to $128 \times 128$ and then randomly cropping it to $112 \times 112$, random horizontal flip, random rotation, and random affine. The model uses a Sub-center ArcFace loss to train the proposed solution.

\vspace{-2mm}
\paragraph{Mask aware ArcFace (MaskedArcFace)} opts to generate a masked twin dataset from MS1MV2 \cite{DBLP:conf/eccv/GuoZHHG16,DBLP:conf/cvpr/DengGXZ19} dataset and to combine them during the training process. Both datasets are shuffled separately using the same seed and, for every new face image selected for the input batch, MaskedArcFace decides whether the image is taken from the original (not-masked) or the masked dataset with a probability of 50\%. MaskedArcFace use ArcFace \cite{DBLP:conf/cvpr/DengGXZ19} as the baseline work. 
%for two reasons: it uses a softmax-loss-based methodology, which does not require an exhaustive training-data-preparation stage, and it has been proven to be the approach that reports excellent results for the original face recognition task. Thus,
MaskedArcFace selects the dataset recommended  by ArcFace (MS1MV2) \cite{DBLP:conf/eccv/GuoZHHG16,DBLP:conf/cvpr/DengGXZ19} as the training dataset, which contains 5.8M images and 85,000 identities.
MaskedArcFace uses IResNet-50 as the backbone among all the network architectures tested in the ArcFace repository as it is it offers good trade-off between the accuracy and the number of parameters. 
%MaskedArcFace uses the Pytorch implementation publicly available in the ArcFace repository.
For the generation of the masked version of the dataset, MaskedArcFace uses MaskTheFace \cite{anwar2020masked}. The types of masks considered are surgical, surgical green, surgical blue, N95, cloth, and KN95. The mask type is selected randomly with a 50\% probability of applying a random color and a 50\% probability of applying a random texture. 
During the evaluation phase, MaskTheFace uses the provided landmark points and the bounding box provided by the competition to align and crop face images. The feature embedding produced by MaskedArcFace solution is of the size 512-D and the input face image is of the size $112 \times 112$ pixels.

\vspace{-2mm}
\paragraph{Multi-task ArcFace  (MTArcFace)} utilized the same training dataset, loss function, backbone, and mask generation method as in MaskedArcFace.  MTArcFace adds another dense layer in parallel to the one used to generate the feature vector by IResNet-50, just after the dropout layer. The new dense layer generates an output with two floats, which correspond to the scores related to the probability that the face is masked or not, respectively. This way, MTArcFace aims to force the network to learn when a face is wearing a mask. This information will also be used by the layer that generates the feature vector. The data preprocessing steps and the size of the feature embedding are identical to the MaskedArcFace.

\vspace{-2mm}
\paragraph{Masked face recognition using non-masked region extraction and fine-tuned recognition model (MFR-NMRE-F)}
Based on the 5-point face landmark detections, the proposed approach identifies a crop that corresponds to the upper facial region where masks are not visible. Then, MFR-NMRE-F fine-tuned a VGG2-SE-ResNet-50 face recognition model for the classification task on these crops using the VGGFace2 \cite{cao2018vggface2} training dataset processed with the RetinaFace \cite{retinaface} detector. For the evaluation, MFR-NMRE-F uses the provided face landmarks provided by MFRC-21, since they correspond closely to the RetinaFace results obtained on the training dataset. Using the landmark coordinates, the MFR-NMRE-F solution extracts the upper face region, extracts feature vectors using the fine-tuned VGG2-SE-ResNet-50 model, and compares features using the cosine similarity measure.
The proposed method is trained using cross-entropy (CE) loss. The input size of the proposed model is $96 \times 192$ and the feature embedding size is 2048-D.
\vspace{-2mm}
\paragraph{Masked face recognition using non-masked region extraction and pre-trained recognition model (MFR-NMRE-B)}   identifies a crop that corresponds to the upper facial region where masks are not visible based on the 5-point face landmark. MFR-NMRE-B utilized a VGG2-SE-ResNet-50  model pre-trained for the classification task using the VGGFace2 \cite{cao2018vggface2} training dataset.
Different from MFR-NMRE-F, the MFR-NMRE-B solution did not fine-tune the feature extraction model with cropped images.
For the evaluation, the proposed method uses the provided face landmarks provided by MFRC-21. Using the landmark coordinates,  the proposed method crops the upper face region, extracts feature vectors using the VGG2-SE-ResNet-50 model, and compares features using the cosine similarity measure. MFR-NMRE-B is trained using Softmax cross-entropy loss. The input size of the proposed model is $112 \times 224$ and the feature embedding size is 2048-D.

\vspace{-2mm}
\paragraph{Masked-Unmasked Face Matching Net (MUFM Net)} utilizes Momentum Contrast (MoCo) \cite{DBLP:conf/cvpr/He0WXG20} to create an initial embedding using a ResNet-50 model trained on CelebA dataset \cite{DBLP:conf/iccv/LiuLWT15}. Then, synthetic masked versions of CelebA, Spectacles on Faces \cite{DBLP:journals/jvcir/AfifiA19}, Youtube Faces \cite{wolf2011face} and LFW \cite{LFWTech} are created
%using the mask generation method 
as defined in \cite{ngan2020ongoing}. The initial model is fine-tuned using these dataset.
For fine-tuning, MUFM Net uses a siamese network with shared weights with absolute differences taken at the last bottleneck layer.
%of the ResNet-50. 
This difference is fed into a 512 fully connected layer followed by a single softmax node.% which outputs the similarity during training. 
%During the inference phase, euclidean distance is obtained from the last layer (2048-D) and converted to similarity.
The model is fine-tuned with binary cross-entropy loss with 50\% of layers frozen. 
%Training data is drawn at random by first drawing a reference ID and then a probe ID with a balanced split of genuine and imposter pairs. During the training phase, random resized crop, and random horizontal flip are used as data augmentation.
%Validation was done using a precision metric.
% every 100 iterations (1 epoch). 
%From the validation set, 20 masked and unmasked pairs are drawn at random - 19 imposter pairs and 1 genuine pair. 
%Evaluation on 20 such pairs counts as one validation iteration. 400 such iterations are conducted each training epoch. 
%Precision over the iteration is counted as the percentage of iterations where the genuine pair has the highest similarity. Training epochs that produce a checkpoint with at least 90\% validation precision were chosen for further evaluation on the testing datasets. 
%The training is stopped after 695k training iterations.
The input size of the presented model is $224 \times 224$ pixels.

\vspace{-2mm}
\paragraph{Ensemble MUFM Net (EMUFM Net)} builds upon MUFM to create an ensemble. 
First, the best-performing MUFM models are selected based on the validation accuracy. The selected models are M1 (obtained after 695K iterations) and M2 (obtained after 885K iteration)
%where each training epoch is counted as 250 training iterations.
These models are fine-tuned on hard examples drawn from the training set.  
%Different from MUFM Net, the identity selection process is modified as follows: instead of drawing a dataset proportional to its size, the EMUFM solution selects the dataset uniformly at random and draws the reference identity uniformly at random from the selected dataset. This allows a more uniform representation of each dataset. 
Three models are fine-tuned- E1 and E2 builds on M1 where 90\% and 80\% of the layers are frozen, respectively, and E3 builds on M2 where  50\% of the layers are frozen.
%with a batch size of 32. 
%E1 builds off M1. This model is fine-tuned by freezing 90\% of the layers and fine-tuning the model with a learning rate of 0.001 for 11k training iterations. The number of hardest imposter pairs for the E1 model is 16. E2 builds off M1 and has 80\% of the layers frozen with a learning rate of 0.01 and the number of hardest imposter pairs is 32. The E2 model is trained for 11250 iterations. E3 builds off M2 and has 50\% of the layers frozen with a learning rate of 0.01 and the number of hardest imposter pairs equals 10 and it is trained for 14500 iterations.
All these models have an input of size $224 \times 224$ and an output embedding of size 2048-D.
During the testing phase, the similarity scores of these three models (E1-3) are averaged to provide the final similarity score. 
%The individual similarity is calculated identically to MUFM by calculating the euclidean distance of the outputs layer (2048-D) and converting it to similarity as $\frac{1}{1 + Distance}$.

\vspace{-2mm}
\paragraph{VIPLFACE-M} adopted ResNet-100 \cite{DBLP:conf/cvpr/HeZRS16} and ArcFace loss \cite{DBLP:conf/cvpr/DengGXZ19} for face recognition. The proposed solution uses a refined version of MS1M dataset \cite{DBLP:conf/eccv/GuoZHHG16} for training the proposed solution. The number of face images in the training dataset is 3.8M of 50K identities. VIPLFACE-M uses the synthetic mask creation method defined in \footnote{\url{https://github.com/JDAI-CV/FaceX-Zoo/blob/main/addition_module/face_mask_adding/FMA-3D/README.md}} to add synthetic masks on part of the training dataset. The number of synthetically masked face images used in the training is 500K and the number of synthetically masked identities is 50K. During the training phase, the proposed solution uses random flipping as a data augmentation method. 
The input size of the presented solution is $ 112 \times 112$ and the output feature embedding size is 512-D.
\vspace{-2mm}
\paragraph{VIPLFACE-G} is based on training ResNet-100 model \cite{DBLP:conf/cvpr/HeZRS16} with ArcFace loss \cite{DBLP:conf/cvpr/DengGXZ19}. The input size of the presented solution is $ 112 \times 112$ and the feature embedding size is 512-D. The model is trained on a clean version of MS1M \cite{DBLP:conf/eccv/GuoZHHG16} that contains 5.8M of 80K identities. The presented solution uses random flip to augment the dataset during training.

\vspace{-2mm}
\paragraph{SiMiT Lab – Masked Face Recognition–1 (SMT-MFR-1)} employs LResNet-100E-IR model \cite{DBLP:conf/cvpr/HeZRS16} trained with ArcFace loss function \cite{DBLP:conf/cvpr/DengGXZ19}. The model is originally trained on MS1MV2 dataset \cite{DBLP:conf/eccv/GuoZHHG16,DBLP:conf/cvpr/DengGXZ19}. SMT-MFR-1 solution depends on fine-tuning LResNet100E-IR using two real world masked face datasets- Real World Occluded Faces (ROF)  \footnotemark\footnotetext{\url{https://github.com/ekremerakin/RealWorldOccludedFaces}} and MFR2 dataset \cite{anwar2020masked}. MFR2 contains 296 images of 53 identities.
ROF dataset is crawled from the internet and contains 678 masked face images and 1853 not-masked face images of 123 identities. The proposed solution is fine-tuned using the ROF dataset and a part of the MFR2 dataset (35 identities). The model process input image of size $112 \times 112$ to produce feature embedding of size 512-D. During the training, the training dataset is augmented using a horizontal flip augmentation method.
\vspace{-2mm}
\paragraph{SiMiT Lab – Masked Face Recognition–2 (SMT-MFR-2)} is conceptually identical to SMT-MFR-1. Different from SMT-MFR-1, the SMT-MFR-2 model is fine-tuned using the ROF dataset and the entire MFR2 dataset.

\vspace{-2mm}
\paragraph{LMI - SiMiT Lab - Masked Face Recognition - 1 (LMI-SMT-MFR-1)} is a combination of two solutions- MFR-NMRE-F and SMT-MFR-1. First, the features are extracted separately by each of the solutions- MFR-NMRE-F and SMT-MFR-1. Then, the comparison scores are calculated for each solution. To combine the scores, cosine similarity measures are converted to euclidean distance in MFR-NMRE-F. The output of SMT-MFR-2 is euclidean distance. After this, the scores are normalized separately for each solution. Then, both scores are multiplied to generate the ensemble score.

\vspace{-2mm}
\paragraph{LMI - SiMiT Lab - Masked Face Recognition - 2 (LMI-SMT-MFR-2)} is also a combination of two solutions-MFR-NMRE-B and SMT-MFR-1. LMI-SMT-MFR-2 follows the same scores fusion method described in the LMI-SMT-MFR-1 solution.

\vspace{-2mm}
\paragraph{Ignoring masks for accurate masked face recognition (IM-MFR)} approach consists of two different training processes. The first, which aims to build a classification model, uses 6000 training identities from the VGGFace2 dataset \cite{cao2018vggface2} to minimize the cross-entropy while classifying these images. Each image had a probability of 65\% of being masked. All training images are randomly resized and cropped to $224 \times 224$
In this solution, the masked creation method \cite{ngan2020ongoing}
uses the open implementation \footnotemark\footnotetext{\url{https://github.com/fdbtrs/MFR/blob/master/FaceMasked.py}} by Boutros \etal \cite{DBLP:journals/corr/abs-2103-01716}. 
%For the validation, only masked images were used. 
%Moreover, for the training, an image was resized to 256x256 and afterward randomly cropped to 224x224. 
After achieving above 96\% accuracy in the classification on the validation set, the last fully-connected layer was replaced with a fully connected layer with 512 outputs units. All the layers, except the newest one, are now frozen. The last layer is trained with and joint Triplet Loss and MSE for metric learning. 
%triplet loss where the anchor is unmasked and both the positive and the negative samples are masked. Furthermore, we also added Mean Squared Error (MSE) loss to minimize the distance between the embedding produced from the masked anchor and the unmasked version of the anchor. %A batch size with 512 triplets was used.
The backbone network is a ResNet-50 \cite{he2016deep}. 
%The suggested approach uses Cross-entropy (CE) loss for the classification loss, and joint Triplet Loss and MSE for metric learning. 
The model is trained for 65k iterations. % with a batch size of 200.
\vspace{-2mm}
\paragraph{Ignoring masks for accurate masked face recognition (IM-AMFR)} follows the same training procedure, architecture, and loss function as in IM-MFR. The only difference is the number of training iteration where the IM-AMFR model is trained for 32k training iterations.
\vspace{-2mm}
\paragraph{anonymous-1 and anonymous-2} employed FaceNet \cite{DBLP:conf/cvpr/SchroffKP15} as base architecture pre-trained on VGGFace2 \cite{cao2018vggface2}. MaskTheFace \cite{anwar2020masked} is used to augment the LFW \cite{LFWTech,DBLP:conf/nips/HuangMLL12} dataset and create a masked-face dataset. 
%MaskTheFace detects facial landmarks using a dlib based detector to identify the list of the face and landmarks of the face required to apply a mask. 
%It then selects a mask from a library of available mask templates. 
A masked version of each image in LFW is created.
%, thus creating a dataset with both masked and unmasked images. 
The FaceNet model is then fine-tuned using the augmented dataset. In the anonymous-1 solution, the model is fine-tuned using only masked face images. In the anonymous-2 solution, the model is fine-tuned using pairs of unmasked and masked images.
%To fine-tune the FaceNet model in the anonymous-2 solution, identical and non-identical pairs from the dataset are created. In total, 100K pairs are created comprising 50k both for identical and non-identical pairs.  Then FaceNet is fine-tuned by dropping the last layers of FaceNet used for VGGFace2 and added two fully connected layers of dimension 256 and 64 respectively to fine-tune the model on the generated pairs using the Cross-entropy (CE) loss. 
For inference, the last layer of FaceNet consists of 512-dimensional embeddings, while the input size for both solutions is $160 \times 160$ pixels. 
One must note that the presented approach is reasonable, however, the verification accuracy presented in Section \ref{sec:res} is extremely low, which might indicate an implementation error in the submission.

\vspace{-2mm}
\paragraph{Baseline}
The baseline is chosen to put the submitted approaches in perspective of state-of-the-art face recognition model performance. The considered baseline is the ArcFace, which scored state-of-the-art performance on several face recognition evaluation benchmarks such as LFW $99.83\%$ and YTF $ 98.02\%$ by using  Additive Angular Margin loss (ArcFace) to improve the discriminative ability of the face recognition model. We considered ArcFace based on ResNet-100 \cite{DBLP:conf/cvpr/HeZRS16} architecture pretrained on refined version of the MS-Celeb-1M dataset \cite{DBLP:conf/eccv/GuoZHHG16} (MS1MV2).

%Overall description of what was asked from the participants

%Overview of the teams and algorithms with a table 

%small description of each algorithm
\begin{figure}[h]
 \centering
    \includegraphics[width=0.99\linewidth]{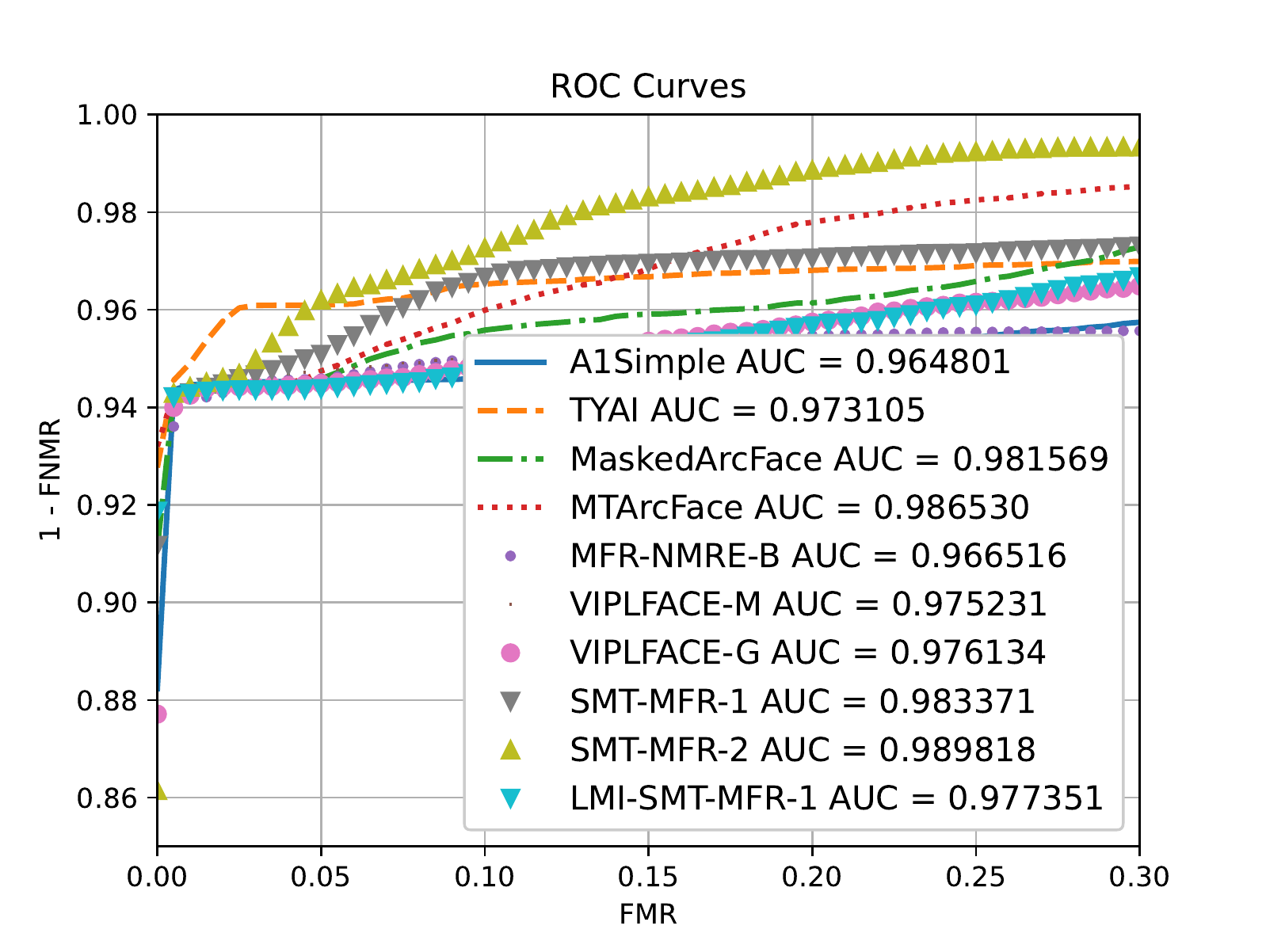}
        \vspace{-4mm}
    \caption{
    The ROC curve scored by the top 10 solutions in the BLR-MP experimental setting.
      }
    \label{fig:roc}
    \vspace{-4mm}
\end{figure}

\begin{figure*}[h]
     \centering
     \begin{subfigure}[b]{0.33\textwidth}
         \centering
         \includegraphics[width=0.99\textwidth]{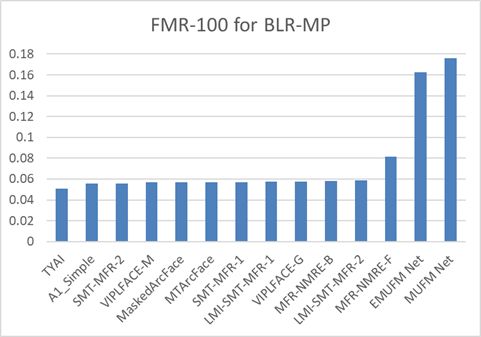}
         \caption{}
         \label{fig:fmr100_BLR_MP}
     \end{subfigure}
     \begin{subfigure}[b]{0.33\textwidth}
         \centering
         \includegraphics[width=0.99\textwidth]{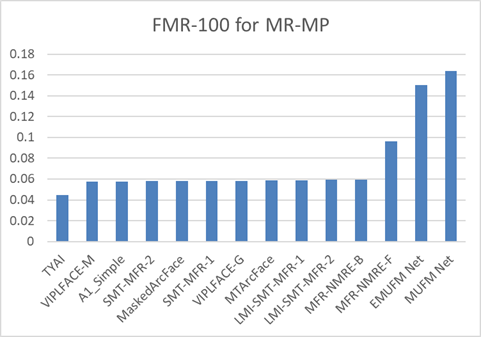}
         \caption{}
         \label{fig:fmr100_MR_MP}
     \end{subfigure}
     \begin{subfigure}[b]{0.33\textwidth}
         \centering
         \includegraphics[width=0.99\textwidth]{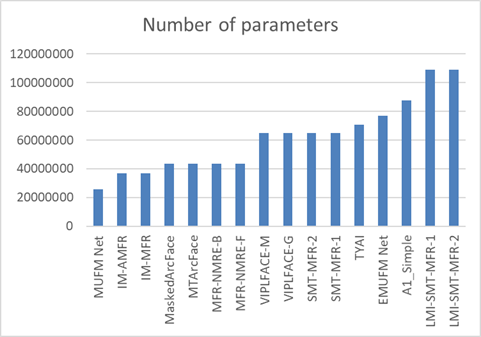}
         \caption{}
         \label{fig:param}
     \end{subfigure}
     \vspace{-4mm}
      \caption{(a) The FMR100 scored by the top 14 solutions in the BLR-MP experimental setting. (b) The FMR100 scored by the top 14 solutions in the MR-MP experimental setting. (c) The number of trainable parameters in the top 16 solutions.}
        \label{fig:vis_comp}
        \vspace{-4mm}
\end{figure*}

\vspace{-2mm}
\section{Results and analyses}
\label{sec:res}
This section presents comparative evaluation results of the submitted solution. We present first the achieved results on the BLR-MP evaluation setting and the model compactness. Then, we present the achieved results on the MR-MP evaluation setting.
\subsection{Not-masked vs. masked (BLR-MP)}
Table \ref{tab:blr_mp} presents comparative evaluation results achieved by the submitted solutions for BLR-MP evaluation setting and the model compactness. The results are reported and ranked based on the evaluation criteria described in Section \ref{sec:eval_criteria}.
%The results are presented in terms of verification evaluation and the model compactness. The verification evaluation is based on the verification performance for the experimental setting BLR-MP including FMR-100, FMR-1000, and FDR. The model compactness is based on the number of parameters. To rank the submitted solution, a) the solutions are ranked first based on the achieved verification performance (FMR-100) b) the Borda count is calculated for each of the submitted solution based on the achieved rank of the verification performance c) the solutions are ranked based on the number of parameters d) the Borda count is calculated for each of the submitted solution based on the model compactness e) a weighted Borda count is calculated based on the compactness rank. The solutions are ordered and ranked based on the final weighted Borda count.
From the reported results in Table \ref{tab:blr_mp} we made the following observations:
\begin{itemize}
    \vspace{-1mm}
    \item Based on the defined evaluation criteria in Section \ref{sec:eval_criteria}, the top-ranked solution based on the weighted Borda count is TYAI (rank 1), followed by MaskedArcFace and SMT-MFR-2 (rank 2) and then A1\_Simple and VIPLFACE-M (rank 3).
        \vspace{-1mm}
    \item Most of the presented solutions achieved a competitive verification performance, in comparison to the baseline. Ten out of 18 solutions achieved higher verification performance than the baseline solution for the BLR-MP evaluation setting as reported in Table \ref{tab:blr_mp} and Figure \ref{fig:roc}. Figure \ref{fig:roc} presented the achieved verification performances in term of Receiver operating characteristic (ROC) curves by the top 10 solution on the BLR-MP experimental setting.
    The best verification performance in terms of FMR100 is achieved by the TYAI solution, where the achieved FMR100 was 0.05095 (Table \ref{tab:blr_mp} and Figure \ref{fig:fmr100_BLR_MP}). 
        \vspace{-1mm}
    \item By comparing the verification performances reported in Table \ref{tab:blr_mp} and the loss function utilized by each of the solution reported in Table \ref{tab:solutions_summary}, it is noted that the models trained with margin-based softmax loss (ArcFace or Sub-center ArcFace loss) achieved higher verification performance than the models trained with other loss functions including cross-entropy and triplet loss. This points out the generalizability brought by the nature of the marginal penalty that forces a better separability between classes (identities) and better compactness within classes.
        \vspace{-1mm}
    \item The solutions that achieved competitive FMR100 to the baseline solution have relatively higher separability between genuine and imposter scores (FDR) than other solutions that achieved relatively lower verification performance.
        \vspace{-1mm}
    \item Regarding model compactness, all solutions contain between 23M and 108M parameters as shown in Table \ref{tab:blr_mp} and Figure \ref{fig:param}. The top 3 ranked solutions have less than 87M parameters. This indicates that utilizing a larger and deeper deep learning model does not necessarily and solely lead to higher verification performance.
        \vspace{-1mm}
    \item The common strategy to improve the masked face recognition verification performance by the submitted solutions is to augment the training dataset with a simulated mask. All submitted solutions depended on training or fine-tuning face recognition model with masked face images (real or simulated). However, none of the presented solutions propose a solution that could be applied on top of the existing face recognition model, as in \cite{DBLP:journals/corr/abs-2103-01716}. Furthermore, none of the presented solutions has clearly benefited from the mask labels included in the evaluation list. 
    Four of the five top-ranked solutions utilized synthetically generate masks to augment the training dataset with simulated masked images. Utilizing such a method is usually easier than other solutions, such as using a real masked training dataset. Collecting a large-scale training dataset with pairs of not-masked/masked face images is, however, not a trivial task. 
\end{itemize}

\vspace{-1mm}
\subsection{Masked vs. masked (MP-MR)}
\vspace{-1mm}
The verification performance of the experimental setting MR-MP for all submitted solutions is presented in Table \ref{tab:mr_mp}. The achieved verification performance is reported in terms of FMR100, FMR1000, and FDR. The presented results are ordered and ranked based on the achieved FMR100. It can be noted from the reported verification performance in Table \ref{tab:mr_mp} that ten out of 18 solutions achieved better verification performance than the baseline solution when comparing masked reference to masked probe (MR-MP).
TYAI solution achieved the best verification performance followed by VIPLFACE-M and A1\_Simple.
By comparing the reported verification performance of BLR-MP evaluation setting (Table \ref{tab:blr_mp}) and the reported one of MR-MP (Table \ref{tab:mr_mp}), we can observe the following: a) Most of the solutions have higher separability between genuine and imposter scores (higher FDR) when both reference and probe are masked (MR-MP) than the case where only the probe are masked (BLR-MP). b) The top-ranked solutions in the MR-MP evaluation setting are also ranked among the top solutions in the BLR-MP evaluation setting.

\section{Conclusion}
\label{sec:con}
Driven by the pandemic-driven use of facial masks, the Masked Face Recognition Competitions (MFR 2021) was organized to motivate and evaluate face recognition solutions specially designed to perform well with masked faces.
A total of 10 teams from 11 affiliations participated in the competition and contributed 18 solutions for the evaluation.
The evaluation focused on not-masked vs. masked face verification accuracy, the masked vs. masked face verification accuracy, and the face recognition model compactness. Out of the 18 submitted solutions, 10 achieved lower verification error (FMR100) than the considered baseline. Most of the top-performing solutions used variations of the ArcFace loss and either real or simulated masked face databases in their training process. The lowest achieved FMR100 for the not-masked vs. masked evaluation was 5.1\%, in comparison to an FMR100 of 6.0\% scored by the baseline.
 
\vspace{-2mm}
{
\paragraph{Acknowledgments:}
This research work has been funded by the German Federal Ministry of Education and Research and the Hessen State Ministry for Higher Education, Research and the Arts within their joint support of the National Research Center for Applied Cybersecurity ATHENE.
}
{\small
\bibliographystyle{ieee}
\bibliography{main}
}

\end{document}